\documentclass[lettersize,journal]{IEEEtran}
\usepackage{amsmath,amsfonts}
\usepackage{algorithmic}
\usepackage{algorithm}
\usepackage{array}
\usepackage[caption=false,font=normalsize,labelfont=sf,textfont=sf]{subfig}
\usepackage{textcomp}
\usepackage{stfloats}
\usepackage{url}
\usepackage{verbatim}
\usepackage{graphicx}
\usepackage{cite}
\usepackage{eccvabbrv}
\usepackage{multirow}
\usepackage{tcolorbox}
\usepackage{resizegather}
\usepackage{colortbl}
\newcommand{\pub}[1]{\color{gray}{\tiny{[{#1}]}}}

\begin{document}
\setlength{\tabcolsep}{3pt} 

\title{GLRD: Global-Local Collaborative Reason and Debate with PSL for 3D Open-Vocabulary Detection}

\author{Xingyu Peng, Si Liu$^\dagger$, Chen Gao, Yan Bai, Beipeng Mu, Xiaofei Wang, Huaxia Xia
        % <-this % stops a space
\thanks{Xingyu Peng, Si Liu, Chen Gao are with the School of Artificial Intelligence, Beihang University, Beijing 100191, China (e-mail: pengxyai@buaa.edu.cn; gaochen.ai@gmail.com; liusi@buaa.edu.cn).}% <-this % stops a space
\thanks{Yan Bai, Beipeng Mu, Xiaofei Wang, Huaxia Xia are with the Meituan, Beijing 100102, China (e-mail: baiyan02@meituan.com; mubeipeng@gmail.com; wangxiaofei19@meituan.com; xiahuaxia@meituan.com).}% <-this % stops a space
\thanks{${^\dagger}$Corresponding Author: Si Liu.}
}

\markboth{Journal of \LaTeX\ Class Files,~Vol.~14, No.~8, August~2021}%
{Shell \MakeLowercase{\textit{et al.}}: A Sample Article Using IEEEtran.cls for IEEE Journals}


\maketitle

\begin{abstract}
The task of LiDAR-based 3D Open-Vocabulary Detection (3D OVD) requires the detector to learn to detect novel objects from point clouds without off-the-shelf training labels. Previous methods focus on the learning of object-level representations and ignore the scene-level information, thus it is hard to distinguish objects with similar classes. 
In this work, we propose a Global-Local Collaborative Reason and Debate with PSL (GLRD) framework for the 3D OVD task, considering both local object-level information and global scene-level information. Specifically, LLM is utilized to perform common sense reasoning based on object-level and scene-level information, where the detection result is refined accordingly. To further boost the LLM's ability of precise decisions, we also design a probabilistic soft logic solver (OV-PSL) to search for the optimal solution, and a debate scheme to confirm the class of confusable objects. 
In addition, to alleviate the uneven distribution of classes, a static balance scheme (SBC) and a dynamic balance scheme (DBC) are designed. In addition, to reduce the influence of noise in data and training, we further propose Reflected Pseudo Labels Generation (RPLG) and Background-Aware Object Localization (BAOL). Extensive experiments conducted on ScanNet and SUN RGB-D demonstrate the superiority of GLRD, where absolute improvements in mean average precision are $+2.82\%$ on SUN RGB-D and $+3.72\%$ on ScanNet in the partial open-vocabulary setting. In the full open-vocabulary setting, the absolute improvements in mean average precision are $+4.03\%$ on ScanNet and $+14.11\%$ on SUN RGB-D.
\end{abstract}


\section{Introduction}
\begin{figure*}[!t]
\centering
\resizebox{\textwidth}{!}{
\includegraphics{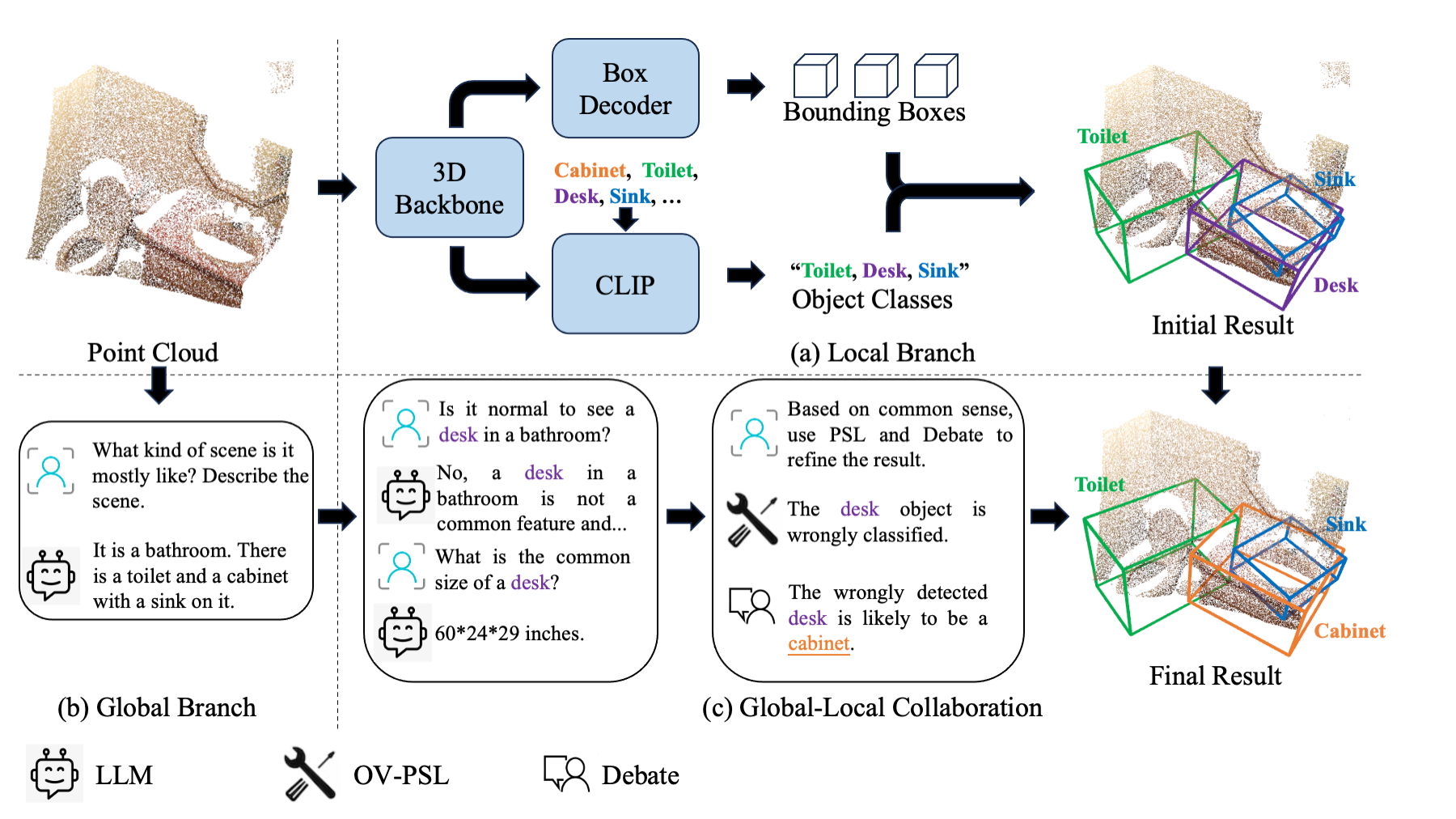}
}
\caption{(a) In LiDAR-Based 3D OVD, the object class may be wrongly recognized when considering only object-level/local information, \eg mistaking the cabinet for a desk. (b)(c) In contrast, GLRD considers both the scene-level/global information and the object-level/local information. Specifically, LLM is utilized to conduct scene understanding and common sense reasoning. Besides to boost LLM's ability of precise decision, a probabilistic soft logic solver and a debate scheme are devised.}
\label{introduction}
\end{figure*}

\IEEEPARstart{R}{ecently}, great success has been witnessed in the field of computer vision~\cite{voulodimos2018deep,guo2022attention,ioannidou2017deep,tang2019object}. As a basic function of computer vision, object detection has been explored in many works, both in the 2D image modal~\cite{chauhan2018convolutional,srivastava2021comparative,ding2021object,ren2016faster} and in the 3D point cloud modal~\cite{qi2019deep, xie2020mlcvnet, ma2021delving,meng2021towards}. Traditional detection model training is dependent on meticulously labeled data, which costs time and effort. In addition, rare classes that are not labeled cannot be detected by a detector trained in such a way. To solve this problem, the Open-Vocabulary Detection (OVD) technique is proposed~\cite{wu2024towards, zhu2024survey}. Basically, OVD supports no labels or labels of a few classes for training and can detect novel classes that do not exist in the off-the-shelf labels. Currently, many efforts have been made to explore OVD in 2D modals~\cite{feng2022promptdet, gupta2022ow, yao2022detclip}. However, the exploration of LiDAR-based 3D Open-Vocabulary Detection (3D OVD) is still limited.

In this paper, we focus on the 3D OVD task, which requires the detector to take the point cloud of a scene as input and output the position and the class of novel objects within the scene. Compared to 2D OVD, 3D OVD faces more challenges. First, point clouds typically possess lower resolutions than 2D RGB images, which results in a loss of object details such as material, texture, and color. Moreover, point cloud quality is susceptible to environmental factors, which can introduce noise into the data. Consequently, 3D OVD models face challenges in recognizing objects when relying solely on object-level information, highlighting the importance of incorporating environmental context in the detection process. 

Therefore, we propose a Global-Local Collaborative Reason and Debate with PSL (GLRD) framework, leveraging the collaboration of local object-level information and global scene-level information to analyze confusable objects detected in 3D point clouds. As shown in Fig.~\ref{introduction}, GLRD is mainly composed of three parts: a local branch to generate initial detection results, a global branch to understand the scene, and a global-local collaborative branch to analyze confusable objects. Primarily, as shown in Fig.~\ref{introduction}(a), with only object-level features, an object is misclassified as a desk. However, with the help of scene-level information and the common sense reasoning of LLM, such an error can be noticed and corrected. Specifically, since the scene is recognized as a bathroom (Fig.~\ref{introduction}(b)), LLM notices that it is not common for a desk to be positioned in a bathroom. In addition, LLM gives the common size of a desk to further confirm the rationality. Taking all these clues as input, a customized probabilistic soft logic solver (OV-PSL) judges that this object is wrongly classified. Finally, the class of this object is corrected through a debate scheme powered by LLM.

Besides, LiDAR-Based 3D OVD suffers from the uneven distribution of classes, especially the novel classes. Such an uneven distribution can cause the detector to focus excessively on certain classes while ignoring other rare classes. To alleviate the uneven distribution of classes, we formulate a static-dynamic balance mechanism, which is composed of a Static Balance between Classes (SBC) module for data balance, and a Dynamic Balance between Classes (DBC) module for training balance. Specifically, SBC automatically adjusts the number of pseudo labels of different classes to achieve balance in data aspect. Taking the loss of different classes in training as evidence, DBC automatically sets the loss weights of different classes to draw more attention to the rare and hard classes.

To facilitate effective global-local collaborative inference, both the local branch and the global branch are carefully designed. In the local branch, we introduce Reflected Pseudo Labels Generation (RPLG), which aims to produce high-quality pseudo labels for training purposes. Furthermore, Background-Aware Object Localization (BAOL) is developed to pick out accurate object proposals. On the other hand, the global branch is trained to perform scene understanding.

In summary, our contributions are as follows:
\begin{itemize}
\item{We introduce a Global-Local Collaborative Reason and Debate with PSL (GLRD) framework, which considers both local object-level information and global scene-level information in the inference of LiDAR-Based 3D Open-Vocabulary Detection. We are the first to explore the interaction of local information and global information in this research field.} 
\item{We devise two tools to boost the LLM’s ability of precise decision in the global-local collaborative inference. Specifically, a probabilistic soft logic solver (OV-PSL) is devised to automatically search for the optimal solution, and a debate scheme is established to confirm the class of confusable object.}
\item{We propose Static Balance between Classes (SBC) to balance the distribution of novel classes in data level. Moreover, Dynamic Balance between Classes (DBC) is introduced to balance the model’s attention towards different classes in training. SBC and DBC form a balance scheme, alleviating the uneven distribution of classes.}
\item{We also propose Reflected Pseudo Labels Generation (RPLG) and Background-Aware Object Localization (BAOL) to reduce the influence of noise.}
\end{itemize}

This work is built upon our conference version~\cite{peng2025global}. We substantially revise and significantly extend the previous work in four aspects. First, we extend the original GLIS to GLRD, significantly revising the pipeline of Global-Local Collaboration. Compared to the original GLIS, GLRD possesses a more comprehensive reasoning framework and takes additional factors (eg., the object size, other common objects in the scene, etc.) into consideration in the global-local collaborative inference. Second, a probabilistic soft logic solver OV-PSL and a debate scheme are devised to boost the LLM's ability of precise decision. Specifically, OV-PSL automatically searches for the optimal solution based on all factors. In addition, the debate scheme is established to confirm the class of confusable objects. Third, SBC and DBC are proposed to achieve balance between classes in both data and training aspects. Specifically, SBC is designed to balance the distribution of novel classes
in data level. Moreover, DBC is introduced to balance the model’s attention towards different classes in the training process. Fourth, while GLIS is only tested in the Full Open-Vocabulary Setting, we extent the experiments of GLRD to the Partial Open-Vocabulary Setting. Besides, more ablation experiments on the proposed methods and more visualizations of the Global-Local Collaboration are presented to demonstrate the effectiveness of our proposed methods. We also add additional technique details and explanations in this work.

\section{Related Work}
\subsection{Open Vocabulary Detection}
\textbf{\textit{2D Open Vocabulary Detection:}} Many works have been done to explore OVD based on 2D RGB images~\cite{feng2022promptdet, gupta2022ow, yao2022detclip, du2022learning, ma2022open, zareian2021open, yao2023detclipv2, rahman2020improved, rahman2020zero, chen2023minigpt, zhong2022regionclip, zang2022open, li2022grounded, liu2023grounding, zhou2022detecting, wu2019detectron2, gao2021room, wang2024visionllm, jhang2024v, huangtraining}. Joint~\cite{rahman2020zero} designs a network to collect semantic concepts from base classes, extracting their features for the localization and classification of novel objects. Ow-detr~\cite{gupta2022ow} proposes an end-to-end transformer-based framework for 2D OVD. Detic~\cite{zhou2022detecting} and OVR-CNN~\cite{zareian2021open} train open-vocabulary detectors with image-text pairs. With the development of image-text pretraining technology, pretrained vision-language models are applied in 2D OVD. For example, HierKD~\cite{ma2022open} transfers knowledge of unseen classes from pretrained vision-language models to detectors by knowledge distillation. DetPro~\cite{du2022learning} learns continuous prompt representations from pretrained vision-language models for the OVD task. Promptdet~\cite{feng2022promptdet} adopts CLIP~\cite{radford2021learning} as detector classifier. RegionCLIP~\cite{zhong2022regionclip} crops regions from images, which are matched with captions by CLIP. Recently, with the rise of Multi-Modal Large Language Models (MLLMs), some MLLMs also possess the ability of OVD. For example, VisionLLM~\cite{wang2024visionllm} models LLM as a decoder for the OVD task. Minigpt-v2~\cite{chen2023minigpt} can conduct various vision tasks with a unified LLM, such as object detection and image caption.

\textbf{\textit{3D Open Vocabulary Detection:}} Recently, 3D OVD is getting more attention in research community~\cite{lu2022open, lu2023open, cao2024coda, zhang2025opensight, zhang2023fm, zhu2023object2scene, wang2024ov, cao2024collaborative, etchegaray2024find, peng2025global,yang2024imov3d,wang2024one,maoopendlign, yao2024open}. OV-3DET~\cite{lu2022open, lu2023open} proposes a de-biased cross-modal contrastive learning mechanism to transfer class information from images to point clouds. CoDA~\cite{cao2024coda, cao2024collaborative} utilizes 3D geometries and 2D OVD semantic priors to assign pseudo labels to novel objects. Besides, CoDA proposes an alignment module, considering both class-agnostic alignments and class-specific alignments. OpenSight~\cite{zhang2025opensight} explores 3D OVD in outdoor scenes. FM-OV3D~\cite{zhang2023fm} extracts knowledge and priors from foundation models (\eg, GPT-3~\cite{brown2020language}, Grounding DINO~\cite{liu2023grounding}, SAM~\cite{kirillov2023segment}, \etal) to support open-vocabulary detection in point clouds. For example, GPT-3 is used to generate descriptions of novel classes, and SAM cuts off specific objects from pictures for contrastive learning. Considering that 3D object datasets cover more classes compared to 3D detection datasets, Object2Scene~\cite{zhu2023object2scene} inserts objects from large-vocabulary 3D object datasets into 3D scenes to enrich the class list of the 3D detection dataset. OV-Uni3DETR~\cite{wang2024ov} proposes Cycle-Modality Propagation to bridge 2D and 3D modalities. Specifically, 2D OVD detectors provide class information for 3D bounding boxes, and trained 3D OVD detectors offer localization supervision for 2D detectors. Find n' Propagate~\cite{etchegaray2024find} designs a strategy to maximize the recall of novel objects. ImOV3D~\cite{yang2024imov3d} generates pseudo point clouds from images and explores joint representations between point clouds and images. OneDet3D~\cite{wang2024one} proposes a joint training pipeline across multi-domain point clouds. 

However, the current dominant paradigm in 3D OVD only considers object-level features, ignoring the value of scene-level information. 

\subsection{LiDAR-based 3D Object Detection}
LiDAR-based 3D Object Detection is critical for many applications, such as autonomous driving, robotics, \etc. Many works have explored the LiDAR-based 3D Object Detection~\cite{guo2020deep,qi2019deep, xie2020mlcvnet, ma2021delving, zhang2020h3dnet, xie2021venet, cheng2021back, shi2020pv, li2021LiDAR, qi2018frustum, shi2019pointrcnn, yang20203dssd, fan2021rangedet, lang2019pointpillars, sun2021rsn, wang2020pillar, yin2021center, zhou2018voxelnet, huang2020epnet, yang2018pixor, chen2017multi, liu2021group, misra2021end, luo20223d, fu2024eliminating}. Some works~\cite{yang2018pixor, chen2017multi} project point clouds to the bird's view to form 2D images, so they can be processed by 2D CNNs. In contrast, PointRCNN~\cite{shi2019pointrcnn} proposes a two-stage framework, which can directly generate 3D proposals from raw point clouds. To improve efficiency, 3DSSD~\cite{yang20203dssd} proposes a single-stage detector for LiDAR-based 3D Object Detection. Unlike directly extracting features from raw point clouds, VoxelNet~\cite{zhou2018voxelnet} partitions the whole point cloud into voxels first and then extracting features for all voxels.  PointPillars~\cite{lang2019pointpillars} accelerates the extraction of point representations. With the rise of Transformer~\cite{vaswani2017attention}, transformer-based 3D detectors also emerge. For example, 3DETR~\cite{misra2021end} designs an end-to-end Transformer-based detector for point clouds, which processes the point clouds using the architecture of the Transformer and extracts object proposals from Transformer queries.

However, these works focus on LiDAR-based 3D Object Detection in close-vocabulary settings. In contrast, our work is towards 3D OVD, the exploration of which is still limited.
\subsection{Large Language Models}
Trained on numerous texts in a self-supervised way, Large Language Models have shown great ability in text generation. For example, GPT-3~\cite{brown2020language} can chat with users and write essays according to instructions. To adapt LLMs to the application of specific areas, many fine-tuning methods are proposed. LoRA~\cite{hu2021lora} injects trainable rank decomposition matrices into LLM layers and freezes pretrained weights to reduce the fine-tuning cost. Unlike directly adjusting the LLMs, P-Tuning~\cite{liu2022p, liu2021p} combines trainable continuous prompt embeddings with discrete prompts, and fulfills the adaption by fine-tuning the trainable prompt embeddings. To enhance LLM's ability of inference, Chain-of-Thought~\cite{wei2022chain} is proposed to guide large models to think step by step. To equip LLMs with the ability to process other modals, \eg, pictures, sounds, \etc, Multi-Modal Large Language Models (MLLMs) are devised. LLaVA~\cite{liu2024visual} connects a visual encoder with an LLM to conduct visual and language understanding. LISA~\cite{lai2024lisa} inserts a special token in the output of the MLLM, which can be decoded into the segmentation mask of the instruction target.

In this paper, we focus on improving the 3D OVD performance with the common sense and inference ability of LLMs.
\subsection{Common sense Reasoning for Visual Understanding}
With the prosperity of machine learning, feature-based visual understanding has made great progress recently. Conventionally, visual inputs (\eg, images, videos, point clouds) are fed into deep neural networks to extract features, and then prediction heads output results accordingly. Although such a feature-based visual understanding pipeline can automatically adapt to different kinds of target objects and has seen great success in applications, it lacks knowledge of the true physical world and cannot distinguish objects from semantic views. To overcome this shortcoming, many works have introduced common sense reasoning into visual understanding tasks~\cite{zellers2019recognition,wang2020visual,zhou2023vicor,gao2023room}. R2C~\cite{zellers2019recognition} proposes the task of Visual Common Sense Reasoning (VCR) and links texts and images with LSTM and attention mechanism. VC R-CNN~\cite{wang2020visual} introduces causal intervention into visual common sense reasoning by inserting actions between different objects. ViCor~\cite{zhou2023vicor} conducts problem classification using LLMs and acquires specific information from images according to the problem type. 

In this paper, we prompt LLMs for common sense reasoning and debate towards the initial detection result from different views, and design a probabilistic soft logic (PSL) ~\cite{bach2017hinge} solver to search for optimal solution. 

\section{Method}
\subsection{Preliminaries}
Basically, a training sample in 3D OVD consists of three components: a 3D point cloud $P$, a 2D RGB image $I$, and a projection matrix $M$. The point cloud $P$ is a set of 3D points $\{(x_i,y_i,z_i)\}_{i=1}^{N_p}$, where $(x_i,y_i,z_i)$ is the 3D coordinate of the i-th point, $N_p$ is the number of points. The 2D RGB image $I\in \mathbb{R}^{H\times W\times 3}$ is in pairs with the point cloud $P$. The projection matrix $M$ is utilized to convert 2D bounding boxes to 3D bounding boxes. Note that in 3D OVD, image $I$ is used only in the training stage, while the input of the testing stage contains only point cloud $P$. The 3D bounding boxes are denoted as $(x,y,z,l,w,h,\theta)$, where $(x,y,z)$ is the center coordinate, $(l,w,h)$ is the length, width, height, respectively, and $\theta$ is the heading angle. A 3D backbone is utilized to extract the local feature $f_{loc} \in\mathbb{R}^{N_{pro}\times D_p}$ and the global feature $f_{glob}\in \mathbb{R}^{1\times D_p}$ from the point cloud $P$, where $N_{pro}$ is the object proposal number of the 3D detector, $D_p$ is the dimension of the 3D feature. 

So far, there are mainly two kinds of settings for evaluating the 3D OVD methods, \ie, the Partial Open-Vocabulary Setting and the Full Open-Vocabulary Setting. In the Partial Open-Vocabulary Setting, object classes are divided into base classes and novel classes. In the training stage, the ground truth labels of the base classes are available, while those of the novel classes are not. Contrastly, the Full Open-Vocabulary Setting is more hard, as no ground truth labels are allowed to use in this setting, \ie, all classes are novel classes. We evaluate our methods in both settings. 
% Besides, since our method is aimed for open-vocabulary detection, the classes mentioned in the following paragraphs are novel classes unless otherwise specified.

\subsection{Overview}
\begin{figure*}[!t]
\centering
\resizebox{\textwidth}{!}{
\includegraphics{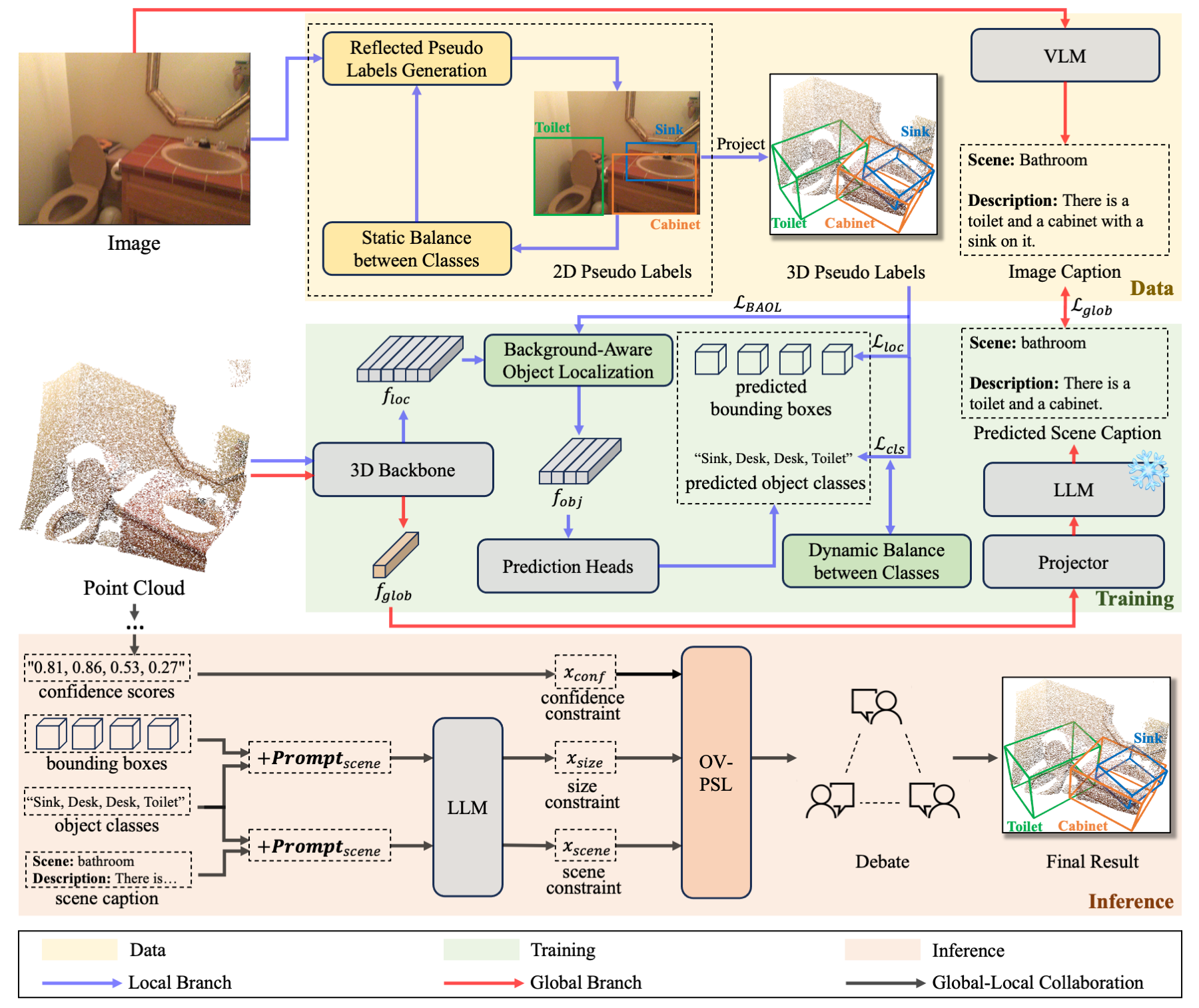}
}
\caption{Overview of GLRD. The GLRD framework enhances 3D open-vocabulary object detection from three aspects: data (yellow block), training (green block), and inference (red block). The yellow block presents the generation of pseudo labels, the green block shows the training pipeline, and the red block demonstrates the workflow of inference. (i) In the data aspect, a circulation is established with Reflected Pseudo Labels Generation (RPLG) and Static Balance between Classes (SBC) to generate precise 3D pseudo labels. (ii) In the training aspect, Background-Aware Object Localization (BAOL) is proposed to distinguish foreground objects from the background and remove low-quality proposals. Besides, Dynamic Balance between Classes (DBC) balances model attention across different classes. (iii) In the inference aspect, LLM is utilized to conduct Global-Local Collaboration and refines the initial detection result. A probabilistic soft logic solver (OV-PSL) is designed to rate scores for each detected object based on common sense constraints.}
\label{overview}
\end{figure*}
\vspace{0.5cm}

The overall framework of our proposed Global-Local Collaborative Reason and Debate with PSL (GLRD) framework is illustrated in Fig. \ref{overview}. GLRD improves the performance of 3D open-vocabulary object detection from three aspects: data, training, and inference. (i) In the data aspect, we formulate a circulation with Reflected Pseudo Labels Generation (RPLG) and Static Balance between Classes (SBC) to generate precise 3D pseudo labels. Besides, a pretrained vision-language model is utilized to conduct image caption. (ii) In the training aspect, the point cloud is input to the 3D Backbone to extract the local feature $f_{loc}$ and the global feature $f_{glob}$. Background-Aware Object Localization (BAOL) further distinguishes foreground objects from the background and removes low-quality proposals, forming $f_{obj}$. Then bounding boxes and object classes are predicted by the prediction heads, which are computed for loss with 3D pseudo labels. Besides, Dynamic Balance between Classes (DBC) is proposed to balance the model's attention across different classes. (iii) In the inference aspect, LLM is utilized to refine the initial detection results in a chain-of-thought pipeline. Firstly, LLM provides common sense information for each object to form constraints. Then a probabilistic soft logic solver (OV-PSL) is utilized to rate scores for each object based on constraints. Based on these scores, the initial detection result is adjusted. Finally, for confusable objects, a debate is conducted using LLM to determine their class. In the following paragraph, we will introduce each proposed module of GLRD in detail.

\subsection{Reflected Pseudo Labels Generation}

\begin{figure}[!t]
\centering
\resizebox{\linewidth}{!}{
\includegraphics{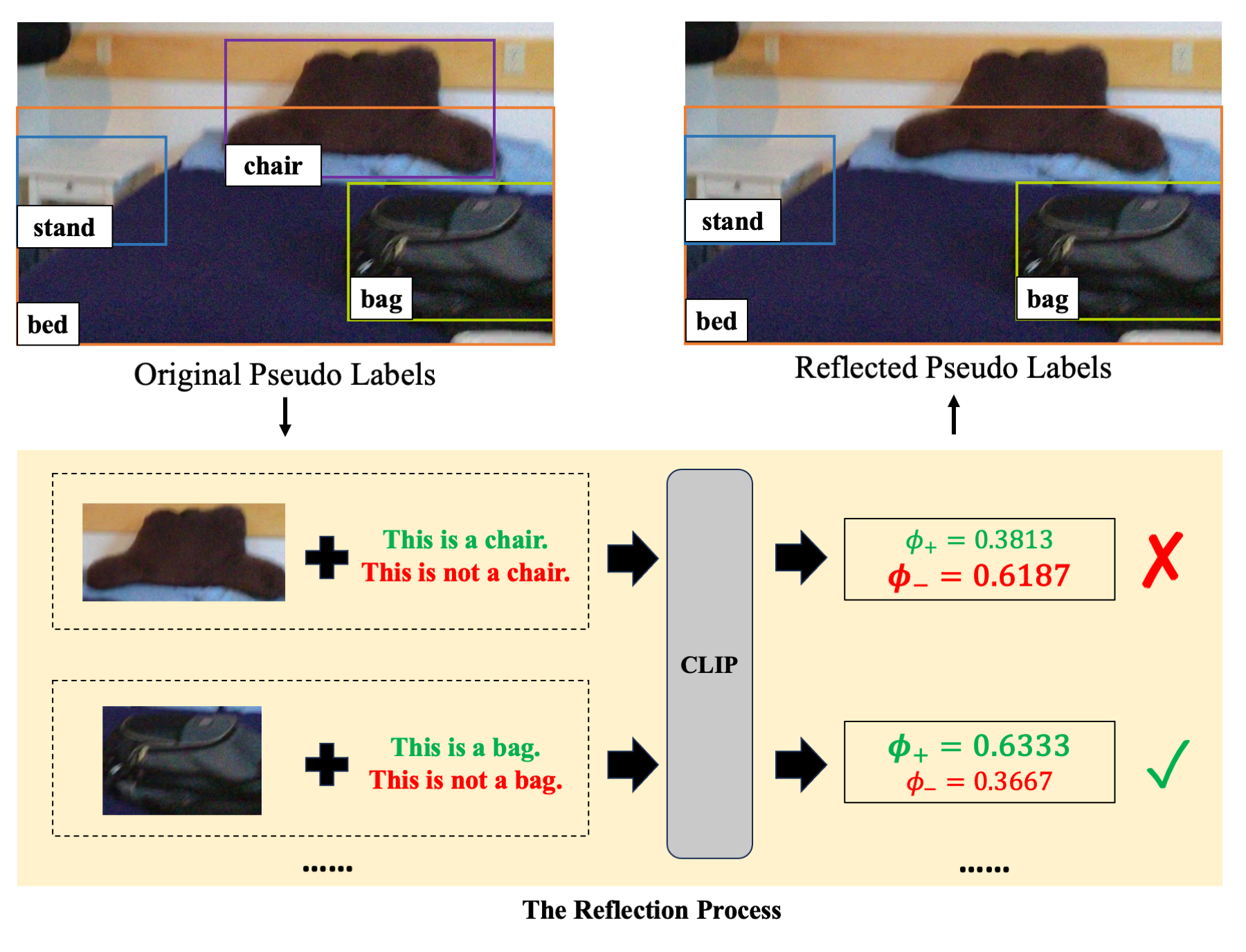}
}
\caption{The Reflected Pseudo Labels Generation (RPLG) module. The image patch and class of each original 2D pseudo label are sent into CLIP with two templates. CLIP judges the image patch's consistency with the class by computing its similarity with two text templates. The labels whose $\phi^+$ is below the threshold $\phi_{CLIP}$ are deleted, forming more accurate pseudo labels.}
\label{RPLG}
\end{figure}
In 3D OVD, no standard labels are available for novel classes. To equip the 3D detector with the ability to detect novel classes, previous methods ~\cite{lu2023open,zhang2023fm,zhang2025opensight,wang2024ov} obtain 2D pseudo labels of novel classes from the corresponding image using 2D open-vocabulary detectors first, and then project these labels onto the point cloud to make 3D pseudo labels. While such a method transfers knowledge from 2D to 3D explicitly, errors (\eg, false detection \& wrong object class) may exist in the generated labels, which can confuse the learning of the 3D detector. To alleviate such confusion, we propose Reflected Pseudo Labels Generation (RPLG), reducing errors in the pseudo labels with a reflection process.

Firstly, original 2D pseudo labels $\{b_{ori}^i, c_{ori}^i\}_{i=1}^{N_{ori}}$ are extracted from the image $I$ by a trained 2D open-vocabulary detector, where $b_{ori}^i$ is the i-th label's 2D bounding box, $c_{ori}^i$ is the i-th label's class, and $N_{ori}$ is the number of original 2D pseudo labels. According to these 2D bounding boxes $\{b_{ori}^i\}_{i=1}^{N_{ori}}$, image patches $\{p_i\}_{i=1}^{N_{ori}}$ are cropped from the image $I$. Now we can form a reflection process with CLIP to check the correctness of each pair $(p_i, c_{ori}^i)$, as shown in Fig. \ref{RPLG}. Specifically, the pair $(p_i, c_{ori}^i)$ is sent into CLIP with text templates $T^+$ and $T^-$:

\begin{center}
$T^+(c_{ori}^i)$: \textit{``This is a \{$c_{ori}^i$\}.",}\\
$T^-(c_{ori}^i)$: \textit{``This is not a \{$c_{ori}^i$\}.".}
\end{center}
Then CLIP computes the similarity of $p_i$ with $T^+$ and $T^-$:

\begin{equation}
[\phi_i^+,\phi_i^-] = \text{Softmax}(\text{CLIP}(T^+(c_{ori}^i),p_i),\text{CLIP}(T^-(c_{ori}^i),p_i)),
\end{equation}

where $\phi_i^+$ is the similarity between $p_i$ and $T^+(c_{ori}^i)$, $\phi_i^-$ is the similarity between $p_i$ and $T^-(c_{ori}^i)$. A higher $\phi_i^+$ means that the class of $p_i$ is more likely to be $c_{ori}^i$. In this way, we delete labels whose $\phi_i^+$ is below the threshold $\phi_{CLIP}$, since its $c_{ori}^i$ does not correspond to the object class in $p_i$. With such a process, origin 2D pseudo labels $\{b_{ori}^i, c_{ori}^i\}_{i=1}^{N_{ori}}$ are transferred into the reflected 2D pseudo labels $\{b_{2d}^i, c_{2d}^i\}_{i=1}^{N_{2d}}$, where $N_{2d}$ is the number of reflected 2D pseudo labels. Apparently, the reflected 2D pseudo labels are more accurate compared to the origin 2D pseudo labels. The reflected 2D pseudo labels $\{b_{2d}^i, c_{2d}^i\}_{i=1}^{N_{2d}}$ are converted to 3D pseudo labels by the projection matrix $M$, which are added to the labels of the base classes. We notate the aggregated 3D labels as $\{b_{3d}^i, c_{3d}^i\}_{i=1}^{N_{3d}}$ where $N_{3d}$ is the total number of 3D labels.

\subsection{Balance Mechanism between Classes}
\begin{figure}[!t]
\centering
\resizebox{\linewidth}{!}{
\includegraphics{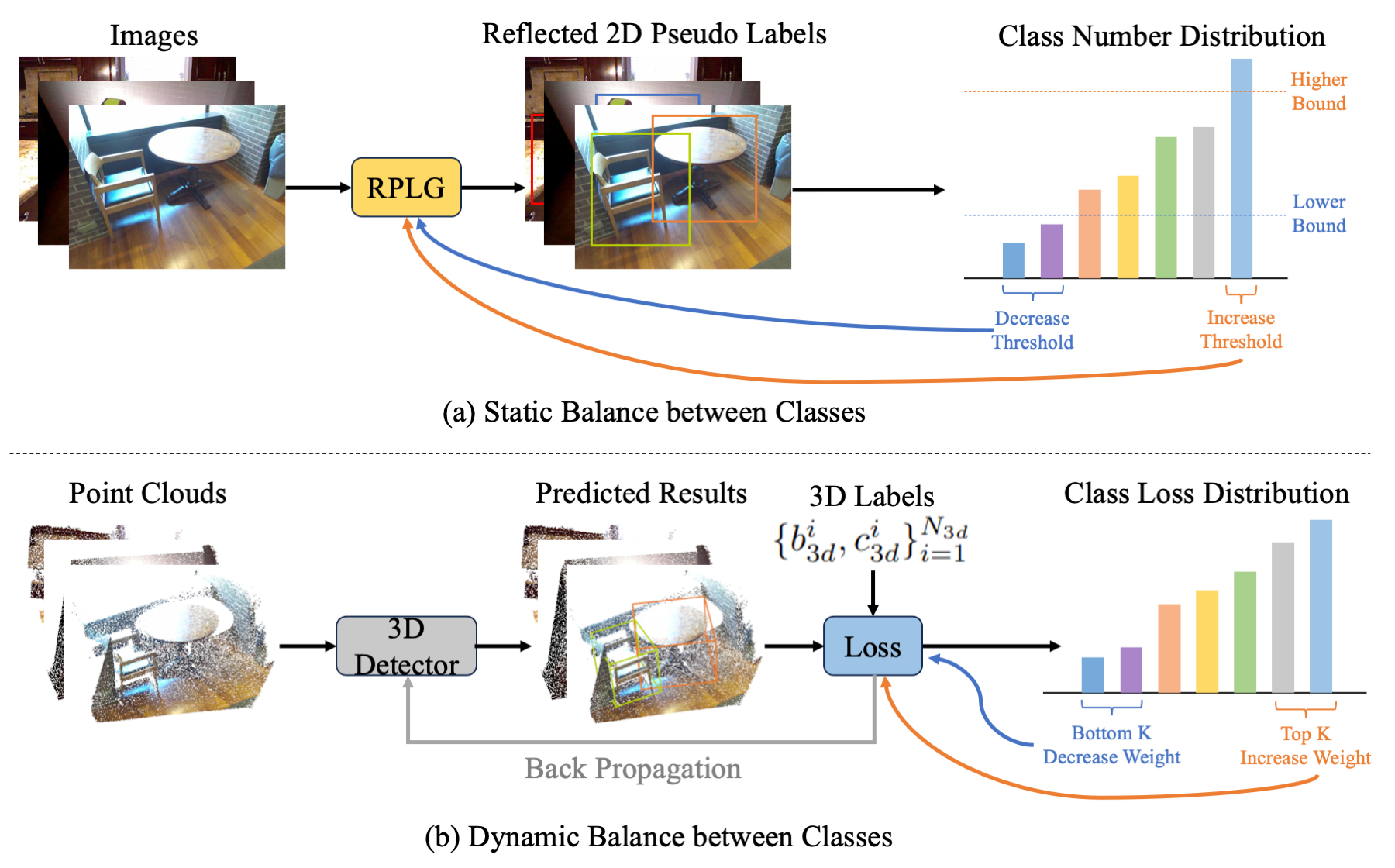}
}
\caption{The balance mechanism is composed of Static Balance between Classes (SBC) and Dynamic Balance between Classes (DBC). (a) SBC balances the number of pseudo labels of different classes by adjusting the confidence threshold automatically. (b) DBC balances the learning efficiency of different classes by adjusting the loss weight automatically.}
\label{balance}
\end{figure}

The RPLG generates 3D pseudo labels from 2D images and provides rich supervisory signals for the 3D detector in training. However, due to the uneven distribution of classes and the limited detection capability of the 2D open-vocabulary detector, the generated pseudo labels can have great difference in number between classes. Furthermore, learning efficiency and difficulty vary from class to class. To deal with the imbalance between classes, we propose a Balance Mechanism between Classes, including Static Balance between Classes (SBC) and Dynamic Balance between Classes (DBC). The SBC adjusts the number of pseudo labels for different classes, while the DBC balances the learning efficiency of different classes according to the training loss. 

\textbf{\textit{Static Balance between Classes (SBC):}} In RPLG, only bounding boxes with confidence greater than a confidence threshold (\ie, $\phi_{2d}$) are output. However, the confidence scores for different classes are not consistent. For example, the confidence scores for the ``nightstand" class are all below $0.5$, while many bounding boxes of the ``picture" class enjoy confidence scores higher than $0.8$. In this way, a unified compromise confidence threshold can lead to a large amount of false detection for the ``picture" class, while the ``nightstand" class lacks pseudo labels at the same time. To solve this problem, we propose the Static Balance between Classes (SBC) to automatically search for an appropriate threshold for each class. The SBC pipeline is illustrated in Fig. \ref{balance}(a). Assume that the confidence threshold for the novel class $c$ is $\phi_c$. In the beginning, all novel classes share the same confidence threshold, \ie, $\phi_c=\phi_{2d}$. After the reflected 2D pseudo labels are generated, the number of pseudo labels for each novel class $c$ is calculated, which is notated as $n_c$. Now we can compute an offset rate $d_c$ for each novel class $c$:
\begin{equation}
n_{avg}=\frac{1}{N_{novel}}\sum_{i=1}^{N_{novel}}n_i,\\
d_c = \frac{n_c-n_{avg}}{n_{avg}},
\end{equation}
where $N_{novel}$ is the number of novel classes. Then confidence threshold for class $c$ is automatically updated as follows:
\begin{equation}
\phi_c\leftarrow\begin{cases}
\phi_c+\text{sgn}(d_c)\cdot \Delta_{\phi},&|d_c|>d_{bound} \land \underline{\phi_{2d}}<\phi_c<\overline{\phi_{2d}},\\
\phi_c,&\text{otherwise},
\end{cases}
\end{equation}
where $\text{sgn}(\cdot)$ is sign function, $\Delta_\phi$ is the step size for each update, $d_{bound}$ is the offset threshold to determine whether $\phi_c$ needs an update, $\underline{\phi_{2d}}$ and $\overline{\phi_{2d}}$ are lower limit and upper limit of $\phi_c$ respectively. With updated confidence thresholds, a new round of pseudo labels generation is conducted, formulating a circulation. This circulation ends when no $\phi_c$ is updated further. Since such a balance mechanism is aimed at static data, we name it as Static Balance between Classes.

\textbf{\textit{Dynamic Balance between Classes (DBC):}} Apart from the imbalance in the number of pseudo labels, the learning efficiency and difficulty of various classes can also differ. Specifically, some classes are easy to learn, whose loss drops and converges quickly in training, while those hard classes hold high training loss all the way. Apparently, the 3D detector should pay more attention to the hard classes as the training progresses. Inspired by this, we designs the Dynamic Balance between Classes (DBC), which is illustrated in Fig. \ref{balance}(b). Firstly, with the predicted results and 3D labels, the original classification loss for class $c$ is calculated, which is recorded as $\mathcal{L}_c$. To adjust the detector's attention towards class $c$, a class weight $w_c$ is utilized to scale $\mathcal{L}_c$:
\begin{equation}
\mathcal{L}'_c=\mathcal{L}_c\cdot w_c,
\end{equation}
where $\mathcal{L}'_c$ is the final classification loss for class $c$. Note that the weight $w_c$ for each class $c$ is initialized as $1$ at the beginning of training, and is updated every $i_{DBC}$ training iterations. To provide evidence for the update of weights, the final classification loss for each class $c$ is accumulated between every two adjacent updates:
\begin{equation}
\mathcal{L}_c^{sum}=\sum_{i=1}^{i_{DBC}}\mathcal{L}'^i_c,
\end{equation}
where $\mathcal{L}'^i_c$ is the final classification loss of class $c$ in the i-th iteration between two weights updates. Then we rank all classes according to $\mathcal{L}'^i_c$. The top-$k$ classes are viewed as hard classes, and their weights increase. In contrast, the bottom-$k$ classes are viewed as easy classes, and their weights decrease. In short, the weight $w_c$ is updated as follows:
\begin{equation}
w_c\gets\begin{cases}
w_c+\Delta_w,&\mathcal{L}_c^{sum}\in \text{top}(\{\mathcal{L}_i^{sum}\}_{i=1}^{N_{class}},k)\land w_c<\overline{w},\\
w_c-\Delta_w,&\mathcal{L}_c^{sum}\in \text{bot}(\{\mathcal{L}_i^{sum}\}_{i=1}^{N_{class}},k)\land w_c>\underline{w},\\
w_c,&\text{otherwise},
\end{cases}
\end{equation}
where $\Delta_w$ is the step size for each update, $\text{top}(\mathcal{A},k)$ returns top-$k$ elements from set $\mathcal{A}$, $\text{bot}(\mathcal{A},k)$ returns bottom-$k$ elements from set $\mathcal{A}$, $N_{class}$ is the number of classes, $k$ is a predefined number, $\overline w$ and $\underline w$ are the upper limit and lower limit we allow $w_c$ to be respectively. These updated weights continue being involved in loss computation, and thus formulate a circulation. Since such a balance mechanism is conducted in the dynamic training process, we call it Dynamic Balance between Classes.

\subsection{Background-Aware Object Localization}
Due to the noise within the point clouds, the 3D detector can confuse the foreground objects with the background noise and detects noisy points as objects falsely. To encourage the 3D detector to distinguish foreground objects from background noise and generate high-quality object proposals, we propose the Background-Aware Object Localization (BAOL). Conventionally, the 3D predictor generates $N_{pro}$ proposals with their possibilities on all classes, forming a class score matrix $S\in\mathbb{R}^{N_{pro}\times N_{class}}$. However, $S$ only focuses on the class of each proposal, which cannot reflect the possibility that the proposal is truly a foreground object instead of background noise. To compensate for this, we utilize a foreground prediction head, which is a liner layer, to predict a foreground score $o_i$ for the i-th proposal. These foreground scores $O=(o_1,\cdots,o_{N_{pro}})$ are used to augment the class score matrix $S$:
\begin{equation}
S_{o}=O^T\circ S,
\end{equation}
where $\circ$ is hadamard product, \ie, the i-th line of $S_{o}$ is the product of $o_i$ and the i-th line of $S$. We pick out top-$k_{pro}$ elements in $S_{o} \in\mathbb{R}^{N_{pro}\times N_{class}}$, and reserve corresponding bounding boxes in $B$:
\begin{equation}
\{(b_i,c_i)\}_{i=1}^{k_{pro}}=\text{arg top}(S_o,k_{pro})\\
B = \text{Set}(\{b_i\}_{i=1}^{k_{pro}})
\end{equation}
where $\text{arg top}(A, k)$ returns the 2D indices of the top-k elements in matrix ${A}$, $\text{Set}(\mathcal A)$ removes duplicate elements from the set $\mathcal A$. In this way, the original class score matrix $S\in \mathbb R^{N_{pro}\times N_{class}}$ is compressed into $S_B \in \mathbb R^{|B| \times N_{class}}$, removing low-quality object proposals. Finally, Soft-NMS~\cite{bodla2017soft} is performed on $S_B$ to obtain the final prediction results.

To supervise the learning of the foreground prediction head, we obtain foreground labels from the aggregated 3D labels $\{b_{3d}^i,c_{3d}^i\}$. Conventionally, the predicted object proposals $\{b_{pro}^i\}_{i=1}^{N_{pro}}$ are matched with the aggregated 3D labels using bipartite matching. The matched proposals are labeled as foreground objects, while other proposals are labeled as background. However, such a straightforward assignment can be inaccurate in two situations: (i) The matched proposal has a low IoU with the matched label, which means this proposal is not accurate enough; (ii) Two different proposals both have high IoU with a label, yet only one of the proposals could be labeled as foreground object, which can cause confusion. In this way, we set thresholds $\underline {\phi_{IoU}}$ and  $\overline {\phi_{IoU}}$, and adjust the assignment of foreground labels as follows: (i) If a matched proposal has an IoU lower than $\underline {\phi_{IoU}}$ with its matched 3D label, it will be seen as background; (ii) If an unmatched proposal has an IoU higher than $\overline {\phi_{IoU}}$, it will be seen as a foreground object. With these adjusted foreground labels, the foreground prediction head can learn to pick out high-quality proposals that are more likely to be foreground objects.

\subsection{Global Scene Understanding}
So far we have finished the local branch learning and obtained initial detection results from the 3D detector. To fulfill global-local collaborative inference, we still need scene-level information from the global branch. Specifically, based on the global feature $f_{glob}$ from the 3D detector, the global branch predicts the scene type $s$ (\eg, bedroom, kitchen, \etc) and subsequently generates a scene description $d$. We achieve this by prompting the LLM with the following text $T_{glob}$:
\begin{center}
\textit{``What kind of scene is it mostly like? Describe the scene.".}
\end{center}
Besides, a global projector is utilized to align the global feature $f_{glob}$ to the LLM embedding space. In short, the whole process could be represented as 
\begin{equation}
s,d = \text{LLM}(T_{glob}, \text{Global-Projector}(f_{glob})).
\end{equation}

To supervise the learning of the global branch, we utilize a pretrained vision-language model to generate scene type labels $\tilde s$ and scene description labels $\tilde d$ from the paired image, which are used for loss computation with the LLM answers.

\subsection{Global-Local Collaboration}
\begin{figure*}[!t]
\centering
\resizebox{\textwidth}{!}{
\includegraphics{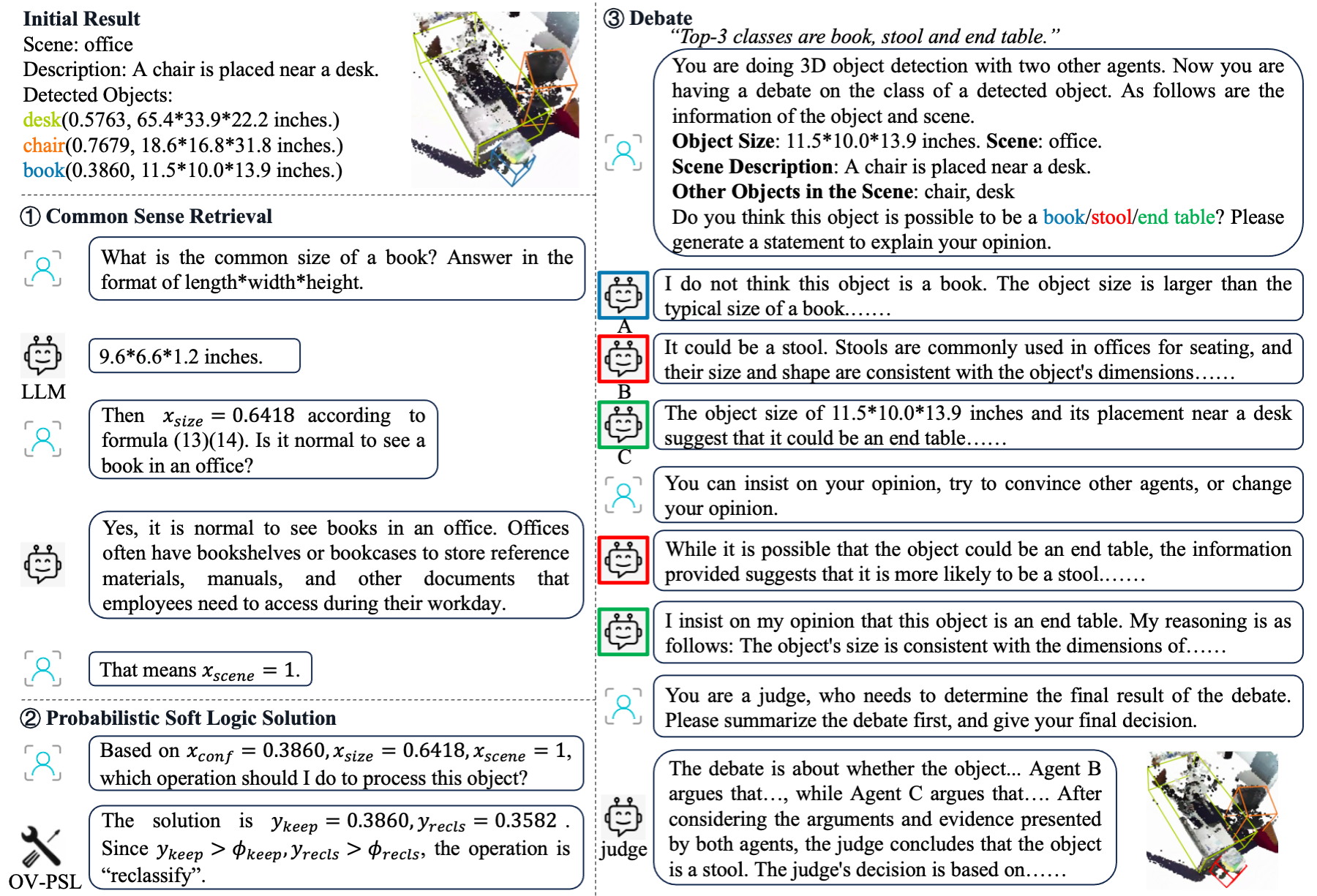}
}
\caption{The pipeline of Global-Local Collaboration. (i) common sense is retrieved from LLM to form constraints $x_{size},x_{scene},x_{conf}$. (ii) The OV-PSL is utilized to work out the optimal operation to the object. The operation is chosen from ``keep/remove/reclassify". (iii) If the operation is ``reclassify", then a debate is conducted to determine the class of the object.}
\label{GLRD}
\end{figure*}

With object-level information from the local branch and scene-level information from the global branch, we can now conduct the Global-Local Collaboration, which is shown in Fig. \ref{GLRD}. Specifically, the detection result from the local branch could be notated as $\{b_i,c_i,s_i\}_{i=1}^{N_{loc}}$, where $b_i$ is the bounding box, $c_i$ is the class, and $s_i$ is the confidence score of the i-th predicted object. The global branch outputs the predicted scene type $s$ and generated scene description $d$. Then we retrieve common sense from LLM to evaluate rationality of detection results from various aspects and output corresponding rationality scores, which we name as confidence constraint, size constraint and scene constraint, respectively. For the i-th object $(b_i,c_i,s_i)$, to evaluate the rationality of its size, we prompt the LLM with following template:
\begin{center}
$\mathbf{Prompt}_{size}(class)$: \textit{``What is the common size of a \{class\}? Answer in the format of length*width*height.".}
\end{center}
In this way, the common size of class $c_i$ can be obtained by:
\begin{equation}
l_{std},w_{std},h_{std} = \text{LLM}(\mathbf{Prompt}_{size}(c_i)),
\end{equation}
where $l_{std}$ is standard length, $w_{std}$ is standard width, and $h_{std}$ is standard height. Then the size constraint $x_{size}$ is given by:
\begin{equation}
x_{size}=\frac{1}{3}[\delta(l,l_{std})+\delta(w,w_{std})+\delta(h,h_{std})],
\end{equation}
where $\delta(x,y)$ is defined as:
\begin{equation}
\delta(x,y)=e^{-\alpha\cdot\max(0,\frac{|x-y|}{y}-\phi_{size})},
\end{equation}
$\alpha$ and $\phi_{size}$ are predefined parameters. Particularly, $\phi_{size}$ is applied to ignore errors that are too small compared to the standard. For the scene constraint $x_{scene}$, we prompt the LLM with the following template to measure the rationality of the object's presence in the scene:
\begin{center}
$\mathbf{Prompt}_{scene}(class,scene)$: \textit{``Is it normal to see a \{class\} in a \{scene\}?".}
\end{center}
The scene constraint $x_{scene}$ will be set as 1 if the LLM judges that the object is reasonable to be present in the scene, or be set as 0 otherwise. The confidence constraint $x_{conf}$ is given the value of the object's confidence score, \ie, $x_{conf}=s_i$. 

Based on above constraints, one of three operations (keep/remove/reclassify) will be performed to the corresponding object. Which operation should be performed is decided by OV-PSL (introduced in Section H), who takes $x_{conf}, x_{size}, x_{scene}$ as input to work out the best decision. If the operation is ``keep", then the object will be kept. If the operation is ``remove", then this object will be deleted from the detection result. If the operation is ``reclassify", then a debate will be conducted to decide its class, as shown in Fig. \ref{GLRD}. Specifically, the top-3 classes with the highest confidence scores for the bounding box are set as choices. Then we ask LLM to play different roles of debaters that support different classes, and utilize a judge role to summarize the debate and give final decision. Note that we only conduct the Global-Local Collaboration for novel classes.

\subsection{Probabilistic Soft Logic Solver for 3D OVD}

To model the complex relationships between the three constraints ($x_{conf},x_{size},x_{scene}$) and operations (keep/remove/reclassify), and work out the final decision, we propose a probabilistic soft logic solver for 3D OVD (OV-PSL), which utilizes probabilistic soft logic~\cite{bach2017hinge} to model the relationships and automatically solves for the decision. Basically, PSL describes the relationships of variables by logical predicates (\eg, $\lor,\land,\lnot,\rightarrow$), and transfers logic expressions into optimization problems, which can be solved automatically. Specifically, for variables $x,y\in[0,1]$, PSL defines operators $\lor,\land,\lnot, \rightarrow$ as follows: 
\begin{equation}
x\land y=\max(x+y-1,0),\\
x\lor y=\min(x+y,1),\\
\lnot x = 1-x,\\
x\rightarrow y =\lnot x \lor y.
\label{def}
\end{equation}
Note that the results of these operations are between 0 and 1, allowing recursive computation. Besides, a higher expression value means that the expression is more likely to be true.  

Now we build the OV-PSL. Specifically, we propose a keeping score $y_{keep}$ and a reclassification score $y_{recls}$, and model their relationships with aforementioned constraints as following expressions:
\begin{equation}
\begin{cases}
L_1:x_{conf}\land x_{size}\land x_{scene}\rightarrow y_{keep}\land\lnot y_{recls},\\
L_2:x_{conf}\land\lnot(x_{size}\land x_{scene})\rightarrow \lnot y_{keep} \lor y_{recls},\\
L_3:\lnot x_{conf}\rightarrow \lnot y_{keep},
\end{cases}
\end{equation}
where $y_{keep}$ decides whether to keep or remove the detected object, and $y_{recls}$ decides whether to change its class. Expression $L_1$ means that if all constraints support that the detected object is correct, it will be kept and its class will not change. Expression $L_2$ means that if a detected object with a high confidence score does not fit its class features (\ie, the size, or the rationality to be in the scene), it will be either removed or reclassified. Expression $L_3$ means that if the detected object's confidence score is low, then it should be removed. With definitions (15)-(18), $L_1,L_2,L_3$ can be transferred to value expressions, and the goal of OV-PSL is to maximize their values. Specifically, the OV-PSL solves the following optimization problem:
\begin{equation}
\begin{aligned}
\mathop{\text{maximize}}\limits_{y_{keep},y_{recls}|\mathbf{x}} &\quad \alpha_1L_1+\alpha_2L_2+\alpha_3L_3,\\
\text{s.t.} &\quad 0\leq y_{keep},y_{recls} \leq 1,
\end{aligned}
\end{equation}

where $\mathbf{x}=(x_{conf},x_{size},x_{scene})$, $\alpha_1,\alpha_2,\alpha_3$ are weights. $y_{keep},y_{recls}|\mathbf{x}$ means that $y_{keep}$ and $y_{recls}$ are optimized on the condition of $\mathbf{x}$, which has been given before. Assuming that the solution is $(y^*_{keep},y^*_{recls})$, we process the corresponding predicted object as follows:

\begin{itemize}
\item{If $y^*_{keep}$ is above the keeping threshold $\phi_{keep}$, then this object will be kept, otherwise, it will be removed.}
\item{If this object is kept and $y^*_{recls}$ is above the reclassification threshold $\phi_{recls}$}, then it will seen as a contested case, whose class will be determined through a debate driven by LLM. Otherwise, it will keep its original class.
\end{itemize}

\subsection{Training Objectives}
The loss for the BAOL module is defined as:
\begin{equation}
\mathcal{L}_{BAOL} = -\frac{1}{N_{pro}}\sum_{i=1}^{N_{pro}}[y_i\log o_i+\lambda_{BAOL}(1-y_i)\log(1-o_i)],
\end{equation}
where $y_i=1$ means that the i-th proposal is labeled as a foreground object, and $y_i=0$ otherwise. $N_{pro}$ is the proposal number, $o_i$ is the predicted possibility that the i-th proposal is a foreground object.

Cross-Entropy loss is utilized to supervise the next token prediction in the global scene understanding. Specifically, assuming the label text is a sequence of tokens $t=(w_1,w_2,\cdots,w_l)$ and the predicted possibility for each token is $p(t)=[p(w_1),p(w_2|w_1),\cdots,p(w_l|w_1,\cdots,w_{l-1})]$, the global scene understanding loss $\mathcal{L}_{glob}$ is defined as:
\begin{equation}
\mathcal{L}_{glob}=-\sum_{i=1}^l \log p(w_i|w_1,\cdots,w_{i-1}).
\end{equation}

We follow OV-Uni3DETR~\cite{wang2024ov} to train the 3D detector.

\section{Experiments}
\subsection{Experimental Setup}
\textbf{\textit{Datasets:}} We evaluate our methods on two datasets: ScanNet~\cite{dai2017scannet} and SUN RGB-D~\cite{song2015sun}. ScanNet is a comprehensive and richly annotated dataset designed for 3D computer vision tasks in the domain of indoor scenes. It consists of 1,513 scans with annotations for more than 200 object categories. By providing instance-level annotations, ScanNet supports various computer vision tasks, such as object detection and instance segmentation. SUN RGB-D dataset is another widely used benchmark in the field of 3D computer vision. It contains 10,335 scenes with about 800 object classes. Apart from 3D bounding boxes and semantic labels, the layout of each scene is also labeled in detail. SUN RGB-D supports computer vision tasks such as depth estimation, object recognition, \etc. 

\textbf{\textit{Settings:}} We conduct comprehensive evaluations in both the Partial Open-Vocabulary Setting and the Full Open-Vocabulary Setting. In the Partial Open-Vocabulary Setting, we follow the benchmarks in Coda~\cite{cao2024coda}. Specifically, for ScanNet, 60 classes are evaluated in the test, where the top 10 classes are base classes (\ie, ground truth labels are available), and the other 50 classes are novel classes. For SUN RGB-D, 46 classes are evaluated in the test, where the top 10 classes are base classes, and the other 36 classes are novel classes. In the Full Open-Vocabulary Setting, no ground truth labels are available in training. Following OV-3DET~\cite{lu2023open}, the top 20 classes are evaluated in the test on both ScanNet and SUN RGB-D. 

\textbf{\textit{Evaluation Metrics:}} Generally, we use the mean Average Precision (mAP) at the IoU threshold of 0.25 for evaluation. In the Partial Open-Vocabulary Setting, we report the mAP of unseen classes, seen classes, and all classes, which are noted as $AP_{25}^{novel}$, $AP_{25}^{base}$ and $AP_{25}^{mean}$ respectively. In the Full Open-Vocabulary Setting, we report the mAP of the top 20 classes following previous methods, which is noted as $AP_{25}^{20cls}$. Besides, to compare with methods that only report the mAP of the top 10 classes, we also report the mAP of our methods for the top 10 classes, which is noted as $AP_{25}^{10cls}$.

\textbf{\textit{Implementation Details:}} The 3D detector is trained for 40 epochs with a learning rate of 2e-4. The BAOL module is trained for 10 epochs with a learning rate of 2e-1. The projector in global scene understanding is trained for 40 epochs with a learning rate of 2e-4. All trainings are completed on 8 A100 GPUs. In RPLG, Detic~\cite{zhou2022detecting} is used to generate 2D labels and $\phi_{CLIP}$ is 0.5. In SBC, $\Delta_{\phi},d_{bound},\underline{\phi_{2d}},\overline{\phi_{2d}}$ are 0.05, 0.5, 0.1, 0.9, respectively. In DBC, $i_{DBC},k,\Delta_w,\overline w,\underline w$ are 2000, 5, 0.05, 0.5, 1.5, respectively. In BAOL, $N_{pro},k_{pro},\underline{\phi_{IoU}},\overline{\phi_{IoU}}$ are 1200, 1000, 0.25, 0.85, respectively. MiniGPT-v2~\cite{chen2023minigpt} is utilized as the vision-language model to generate scene type labels and scene description labels from images. In formula (14), $\alpha, \phi_{size}$ are set to be 0.25, 0.05, respectively. In PSL, $\phi_{keep}$ is 0.01 and $\phi_{recls}$ is 0.2. We choose Uni3detr~\cite{wang2024uni3detr} as the 3D detector, and use LLaMA~\cite{touvron2023llama} with checkpoint vicuna-7b-v1.5-16k~\cite{chiang2023vicuna} as LLM. 

\subsection{Comparison With State-of-the-Art Methods}

\begin{table*}[!t]
\caption{Comparisons with other methods on SUN RGB-D and ScanNet in the Partial Open-Vocabulary Setting}
\centering
\resizebox{\textwidth}{!}{
\begin{tabular}{>{\raggedleft\arraybackslash}p{2.5cm}>{\raggedright\arraybackslash}p{1.0cm}||ccc||ccc}
\hline

 \rowcolor[HTML]{f8f9fa} \multicolumn{2}{c||}{}& \multicolumn{3}{c||}{SUN RGB-D} & \multicolumn{3}{c}{ScanNet}\\
\rowcolor[HTML]{f8f9fa} \multicolumn{2}{c||}{\multirow{-2}{*}{Method}} & $AP_{25}^{novel}$& $AP_{25}^{base}$ & $AP_{25}^{mean}$ & $AP_{25}^{novel}$& $AP_{25}^{base}$ & $AP_{25}^{mean}$\\
\hline
\hline
Det-PointCLIP~\cite{zhang2022pointclip}\!\!&\!\!\pub{CVPR2022} & 0.09 & 5.04 & 1.17 & 0.13 & 2.38 & 0.50\\
Det-PointCLIPv2~\cite{zhu2023pointclip}\!\!&\!\!\pub{ICCV2023} & 0.12 & 4.82 & 1.14 & 0.13 & 1.75 & 0.40\\ 
Det-CLIP$^2$~\cite{zeng2023clip2}\!\!&\!\!\pub{CVPR2023} & 0.88 & 22.74 & 5.63 & 0.14 & 1.76 & 0.40\\
3D-CLIP~\cite{radford2021learning}\!\!&\!\!\pub{ICML2021} & 3.61 & 30.56 & 9.47 & 3.74 & 14.14 & 5.47 \\
CoDA~\cite{cao2024coda}\!\!&\!\!\pub{NIPS2024} & 6.71 & 38.72 & 13.66 & 6.54 & 21.57 & 9.04\\
INHA~\cite{jiao2024unlocking}\!\!&\!\!\pub{ECCV2024} & 8.91 & 42.17 & 16.18 & 7.79 & 25.1 & 10.68 \\
CoDAv2~\cite{cao2024collaborative}\!\!&\!\! & 9.17 & 42.04 & 16.31 & 9.12 & 23.35 & 11.49\\

OV-Uni3DETR~\cite{wang2024ov}\!\!&\!\!\pub{ECCV2024} & 9.66 & 48.29 & 18.06 & 12.09 & \bf{30.47} & 15.15 \\
GLRD~(ours) \!\!&\!\! &\bf{12.96}~(\textcolor[RGB]{230,0,0}{+3.30})& \bf{49.40}~(\textcolor[RGB]{230,0,0}{+1.11})& \bf{20.88}~(\textcolor[RGB]{230,0,0}{+2.82})& \bf{17.29}~(\textcolor[RGB]{230,0,0}{+5.20})& 26.78&\bf{18.87}~(\textcolor[RGB]{230,0,0}{+3.72})\\
\hline
\end{tabular}
}
\label{coda}
\end{table*}

\begin{table*}[!t]
\caption{Comparisons with other methods on ScanNet in the Full Open-Vocabulary Setting}
\centering
\resizebox{\textwidth}{!}{
 \begin{tabular}{>{\raggedleft\arraybackslash}p{2.3cm}>{\raggedright\arraybackslash}p{1.0cm}||c||ccccccccccccccccccccc}
    \hline
    \rowcolor[HTML]{f8f9fa} \multicolumn{2}{c||}{Method}&$AP_{25}^{10cls}$&toilet&bed&chair&sofa&dresser&table&cabinet&bookshelf&pillow&sink\\
    \hline
    \hline
    OV-3DETIC~\cite{lu2022open}\!\!&\!\! & 12.65 &48.99&2.63&7.27&18.64&2.77&14.34&2.35&4.54&3.93&21.08\\
    FM-OV3D~\cite{zhang2023fm}\!\!&\!\!\pub{AAAI2024}&21.53&62.32&41.97&22.24&31.80&1.89&10.73&1.38&0.11&12.26&30.62\\
    OV-3DET~\cite{lu2023open}\!\!&\!\!\pub{CVPR2023}&24.36&57.29&42.26&27.06&31.50&8.21&14.17&2.98&5.56&23.00&31.60\\
    L3Det~\cite{zhu2023object2scene}\!\!&\!\!&24.62&56.34&36.15&16.12&23.02&8.13&23.12&\bf{14.73}&17.27&23.44&27.94\\
    CoDA~\cite{cao2024coda}\!\!&\!\!\pub{NIPS2024}&28.76&68.09&44.04&28.72&44.57&3.41&20.23&5.32&0.03&27.95&45.26\\

    INHA~\cite{jiao2024unlocking}\!\!&\!\!\pub{ECCV2024}&30.06 & 67.40 & 46.01 & 33.32 & 40.92 & 9.1 & 26.42 & 4.28 & 11.30 & 26.15 & 35.69\\
    GLIS~\cite{peng2025global}\!\!&\!\!\pub{ECCV2024}&30.94&73.90&39.69&39.51&44.41&6.09&25.38&5.92&8.31&25.63&43.51\\
    
    CoDAv2~\cite{cao2024collaborative}\!\!&\!\!&30.06 & 77.24 & 43.96 & 15.05 & 53.27 & 11.37 & 19.36 & 1.42 & 0.11 & 34.42 & 44.38\\

    OV-Uni3DETR~\cite{wang2024ov}\!\!&\!\!\pub{ECCV2024}&34.14 &86.05&50.49&28.11&31.51&\bf{18.22}&24.03&6.58&12.17&29.62&54.63&\\
    
    GLRD~(ours) \!\!&\!\!&\bf{41.26}~(\textcolor[RGB]{230,0,0}{+7.12})&\bf{87.40}&\bf{56.92}&\bf{39.87}&\bf{63.87}&9.88&\bf{35.28}&3.44&\bf{20.91}&\bf{39.15}&\bf{55.89}\\
    \hline
\rowcolor[HTML]{f8f9fa} \multicolumn{2}{c||}{Method}&$AP_{25}^{20cls}$&bathtub&refrigerator&desk&nightstand&counter&door&curtain&box&lamp&bag\\
\hline
    \hline
    OV-3DET~\cite{lu2023open}\!\!&\!\!\pub{CVPR2023}&18.02&56.28&10.99&19.72&0.77&0.31&9.59&10.53&3.78&2.11&2.71&\\
    CoDA~\cite{cao2024coda}\!\!&\!\!\pub{NIPS2024}&19.32&50.51&6.55&12.42&15.15&0.68&7.95&0.01&2.94&0.51&2.02\\
    GLIS~\cite{peng2025global}\!\!&\!\!\pub{ECCV2024} &20.83&53.21&4.76&20.79&7.62&0.09&0.95&7.79&3.32&3.73&1.93\\ 
    ImOV3D~\cite{yang2024imov3d}\!\!&\!\!\pub{NIPS2024}& 21.45 & - & - & - & -&-&-&-&-&-&-\\
    CoDAv2~\cite{cao2024collaborative}\!\!&\!\!&22.72 & 55.60 & \bf{24.41} & 20.67 & \bf{20.72} & 0.28 & \bf{13.54} & 0.92 & 4.16 & 4.37 & 9.20\\
    OV-Uni3DETR~\cite{wang2024ov}\!\!&\!\!\pub{ECCV2024}&25.33&63.73&14.41&30.47&2.94&\bf{1.00}&1.02&19.90&\bf{12.70}&5.58&\bf{13.46}\\
    GLRD~(ours) \!\!&\!\!&\bf{29.36}~(\textcolor[RGB]{230,0,0}{+4.03}) &\bf{65.80}&14.38&\bf{31.16}&10.06&0.47&5.46&\bf{31.03}&6.32&\bf{5.86}&4.01\\
    
  \hline
  \end{tabular}
}
\label{ov_scannet}
\end{table*}

\begin{table*}[!t]
\caption{Comparisons with other methods on SUN RGB-D in the Full Open-Vocabulary Setting}
\centering
\resizebox{\textwidth}{!}{
\begin{tabular}{>{\raggedleft\arraybackslash}p{2.3cm}>{\raggedright\arraybackslash}p{1.0cm}||c||ccccccccccccccccccccc}
    \hline
\rowcolor[HTML]{f8f9fa} \multicolumn{2}{c||}{Method}&$mAP_{25}^{10cls}$&~toilet~&~~bed~~&chair&~bathtub~&~~~sofa~~~&dresser&scanner&fridge&~lamp~&desk\\
\hline
\hline
OV-3DETIC~\cite{lu2022open}\!\!&\!\!&13.03&43.97&6.17&0.89&45.75&2.26&8.22&0.02&8.32&0.07&14.60\\
FM-OV3D~\cite{zhang2023fm}\!\!&\!\!\pub{AAAI2024}&21.47&55.00&38.80&19.20&41.91&23.82&3.52&0.36&5.95&17.40&8.77\\
L3Det~\cite{zhu2023object2scene}\!\!&\!\!&25.42&34.34&54.31&29.84&51.65&34.12&17.12&\bf{5.23}&13.87&11.40&15.32\\
OV-3DET~\cite{lu2023open}\!\!&\!\!\pub{CVPR2023}&31.06&72.64&66.13&34.80&44.74&42.10&11.52&0.29&12.57&14.64&11.21\\
GLIS~\cite{peng2025global}\!\!&\!\!\pub{ECCV2024}&30.83&69.88&63.83&34.78&49.62&40.78&10.73&1.49&8.37&16.40&12.44&\\
GLRD~(ours)\!\!&\!\! & \bf{48.20}~(\textcolor[RGB]{230,0,0}{+17.37})&\bf{89.87}&\bf{85.43}&\bf{72.24}&\bf{70.45}&\bf{66.43}&\bf{18.28}&1.04&\bf{20.62}&\bf{33.00}&\bf{24.67}\\
\hline
\rowcolor[HTML]{f8f9fa} \multicolumn{2}{c||}{Method}&$mAP_{25}^{20cls}$&~table~ &~~stand~~&cabinet&~counter~&~~~bin~~~&bookshelf&pillow&microwave&~sink~&stool\\
\hline
\hline
OV-3DET~\cite{lu2023open}\!\!&\!\!\pub{CVPR2023}&20.46&23.31&2.75&3.40&0.75&23.52&9.83&10.27&1.98&18.57&4.10\\
GLIS~\cite{peng2025global}\!\!&\!\!\pub{ECCV2024}&21.45&19.17&13.84&2.75&0.59&22.22&12.65&15.78&5.30&27.62&0.84\\
ImOV3D~\cite{yang2024imov3d}\!\!&\!\!\pub{NIPS2024}& 22.53 & - & - & - & -&-&-&-&-&-&-\\
GLRD~(ours)\!\!&\!\!& \bf{36.64}~(\textcolor[RGB]{230,0,0}{+14.11})&\bf{48.20}&\bf{35.36}&\bf{4.12}&\bf{0.81}&\bf{39.17}&\bf{15.81}&\bf{25.11}&\bf{15.09}&\bf{53.43}&\bf{13.62}\\

  \hline
  \end{tabular}
}
\label{ov_sunrgbd}
\end{table*}

\textbf{\textit{Partial Open-Vocabulary Setting:}} The results in the Partial Open-Vocabulary Setting ares shown in the Tab. \ref{coda}. GLRD greatly improves the detection precision of novel classes on both datasets, \ie, $A_{25}^{novel}$ is improved by $3.3\%$ on SUN RGB-D and $5.2\%$ on ScanNet, demonstrating the superiority of GLRD in detecting novel objects in 3D scenes. Besides, GLRD also achieves state-of-the-art performance on mean average precision, \ie, $2.82\%$ improvement on SUN RGB-D, $3.72\%$ improvement on ScanNet.

\textbf{\textit{Full Open-Vocabulary Setting:}} The detection precisions in full open-vocabulary setting are shown in Tab. \ref{ov_scannet} and \ref{ov_sunrgbd}. As shown in the Tab. \ref{ov_scannet}, GLRD improves $AP_{25}^{10cls}$ by $7.12\%$ and $AP_{25}^{20cls}$ by $4.03\%$ on ScanNet. Besides, GLRD achieves state-of-art-detection performance of 19 classes on SUN RGB-D, as shown in \ref{ov_sunrgbd}. Such results demonstrate that GLRD still performs well even if no base classes exist. 

\subsection{Ablation Studies}
\begin{table}[!t]
\caption{Ablation studies on SUN RGB-D in the Partial Open-Vocabulary Setting}
\centering
\resizebox{\linewidth}{!}{
\begin{tabular}{c||ccc}
    \hline
\rowcolor[HTML]{f8f9fa} Module&$AP_{25}^{novel}$&$AP_{25}^{base}$&$AP_{25}^{mean}$\\
\hline
\hline
Baseline &9.66&48.29&18.06\\
+SBC&10.56~(\textcolor[RGB]{230,0,0}{+0.90})&48.67~(\textcolor[RGB]{230,0,0}{+0.38})&18.84~(\textcolor[RGB]{230,0,0}{+0.78})\\
+RPLG&10.67~(\textcolor[RGB]{230,0,0}{+0.11})&48.88~(\textcolor[RGB]{230,0,0}{+0.21})&18.98~(\textcolor[RGB]{230,0,0}{+0.14})\\
+DBC&10.86~(\textcolor[RGB]{230,0,0}{+0.19})&48.89~(\textcolor[RGB]{230,0,0}{+0.01})&19.13~(\textcolor[RGB]{230,0,0}{+0.15})\\
+BAOL&10.95~(\textcolor[RGB]{230,0,0}{+0.09})&49.40~(\textcolor[RGB]{230,0,0}{+0.51})&19.31~(\textcolor[RGB]{230,0,0}{+0.18})\\
+Collab. w/o OV-PSL& 11.78~(\textcolor[RGB]{230,0,0}{+0.83})& 49.40& 19.96~(\textcolor[RGB]{230,0,0}{+0.65})\\
+OV-PSL&12.96~(\textcolor[RGB]{230,0,0}{+1.18})&49.40&20.88~(\textcolor[RGB]{230,0,0}{+0.92})\\

  \hline
  \end{tabular}
}
\label{ablation}
\end{table}

The ablation results on SUN RGB-D in the Partial Open-Vocabulary Setting is presented in Tab. \ref{ablation}.

\textbf{\textit{Baseline:}} We choose OV-Uni3DETR~\cite{wang2024ov} as our baseline.

\begin{figure}[!t]
\centering
\resizebox{\linewidth}{!}{
\includegraphics{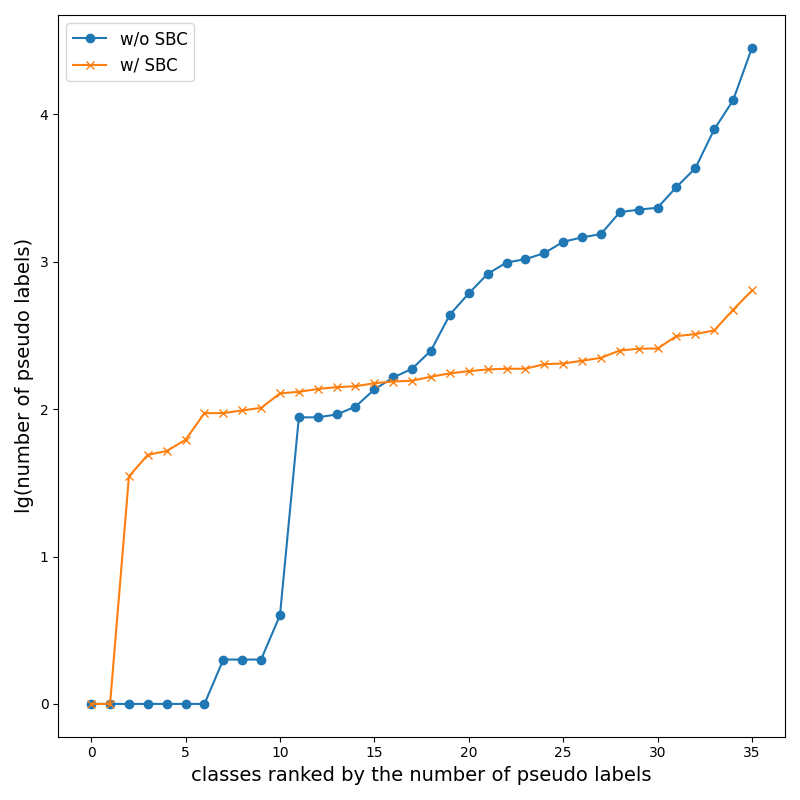}
}
\caption{The number of pseudo labels of novel classes with/without SBC on SUN RGB-D in Partial Open-Vocabulary Setting. With SBC, the number of pseudo labels for novel classes are more balanced.}
\label{abla_SBC}
\end{figure}

\textbf{\textit{Static Balance between Classes (SBC):}} SBC greatly improves the detection performance of novel classes, which increases from $9.66\%$ to $10.56\%$. It should be noted that many rare classes are better detected as more labels of these classes are involved in training, \eg, nightstand is improved from $0.06\%$ to $5.82\%$, bookshelf is improved from $5.19\%$ to $13.56\%$, \etc. To further show the effect of SBC, we count the pseudo labels of the novel classes with/without SBC in Fig. \ref{abla_SBC}, which shows that the number of pseudo labels for novel classes are more balanced with SBC.

\textbf{\textit{Reflected Pseudo Labels Generation (RPLG):}} When the reflection process is involved in the pseudo labels generation, $AP_{25}^{mean}$ is raised by $0.14\%$. This increment shows that RPLG can detect errors within pseudo labels and improve the quality of the local branch supervision. 

\begin{table}[!t]
\caption{Effect of DBC in the Partial Open-Vocabulary Setting}
\centering
\resizebox{\linewidth}{!}{
\begin{tabular}{c||c|c|c|c}
\hline
\rowcolor[HTML]{f8f9fa} class& DBC &3000 iters & 6000 iters & 9000 iters \\
\hline
\hline
\multirow{3}{*}{coffee table} & w/o & 31.40 & 28.13 & 26.96 \\
\cline{2 - 5} 
 & w/ &  24.46  &  33.14  &  34.82  \\
 & (weight) & (1.00) & (1.05) & (1.00) \\
\hline
\multirow{3}{*}{recycle bin} & w/o & 17.09 & 15.33 & 15.06 \\
\cline{2 - 5} 
 & w/ &  17.18  &  16.40  &  18.68  \\
  & (weight) & (1.05) & (1.05) & (1.05) \\
\hline
\multirow{3}{*}{sofa} & w/o & 58.17 & 66.38 & 69.36 \\
\cline{2 - 5} 
 & w/ &  58.79  &  67.80  &  70.53  \\
  & (weight) & (0.95) & (0.95) & (0.95) \\
\hline
\end{tabular}

}
\label{abla_DBC}
\end{table}
\textbf{\textit{Dynamic Balance between Classes (DBC):}}
It could be seen from the Tab. \ref{ablation} that DBC enhances the detector's ability to detect novel objects. Specifically, $AP_{25}^{novel}$ is improved from $10.67\%$ to $10.86\%$. To further demonstrate the effect of DBC, we present cases in \ref{abla_DBC}. The numbers in parentheses are the class weights of certain training iterations. Without DBC, the detecting performance of coffee table declines with the training process, yet the trained model cannot remedy this situation. Differently, with DBC, the weight of coffee table class is increased from 1.00 to 1.05 between 3000 iterations and 6000 iterations, leading the model to pay more attention to coffee table and increase its detection performance. When the detection of coffee table is approaching stable, the wight reverts to 1.00, showing the flexibility of DBC. For classes whose weights decrease, their detection is not affected by DBC (\eg, sofa). In short, DBC dynamically controls the weights of classes to achieve a more balanced training process.

\textbf{\textit{Background-Aware Object Localization (BAOL):}} With the addition of BAOL, $AP_{25}^{mean}$ is raised by $0.18\%$, demonstrating that BAOL can pick out object proposals of good quality and reduce the influence of background noise.

\begin{table}[!t]
\caption{Ablation of $\alpha$ on SUN RGB-D in the Partial Open-Vocabulary Setting}
\centering
\resizebox{0.65\linewidth}{!}{
\begin{tabular}{c||cc}
    \hline
\rowcolor[HTML]{f8f9fa} $\alpha$&$AP_{25}^{novel}$&$AP_{25}^{mean}$\\
\hline
\hline
0.05 &11.30&19.58\\
0.15 &11.73&19.92\\
0.25 &\bf{12.96}&\bf{20.88}\\
0.35 &12.95&20.87\\
0.45 &9.66&18.06\\

  \hline
  \end{tabular}
}
\label{abla_alpha}
\end{table}

\begin{table}[!t]
\caption{Ablation of $\phi_{size}$ on SUN RGB-D in the Partial Open-Vocabulary Setting}
\centering
\resizebox{0.65\linewidth}{!}{
\begin{tabular}{c||cc}
\hline
\rowcolor[HTML]{f8f9fa} $\phi_{size}$&$AP_{25}^{novel}$&$AP_{25}^{mean}$\\
\hline
\hline
0 & 11.49 & 19.73 \\
0.03 & 12.92 & 20.85 \\
0.05 & \bf{12.96} & \bf{20.88} \\
0.07 &12.06&20.18\\
0.09 &11.81&19.98\\
\hline
\end{tabular}
}
\label{abla_size}
\end{table}

\begin{figure}[!t]
\centering
\resizebox{\linewidth}{!}{
\includegraphics{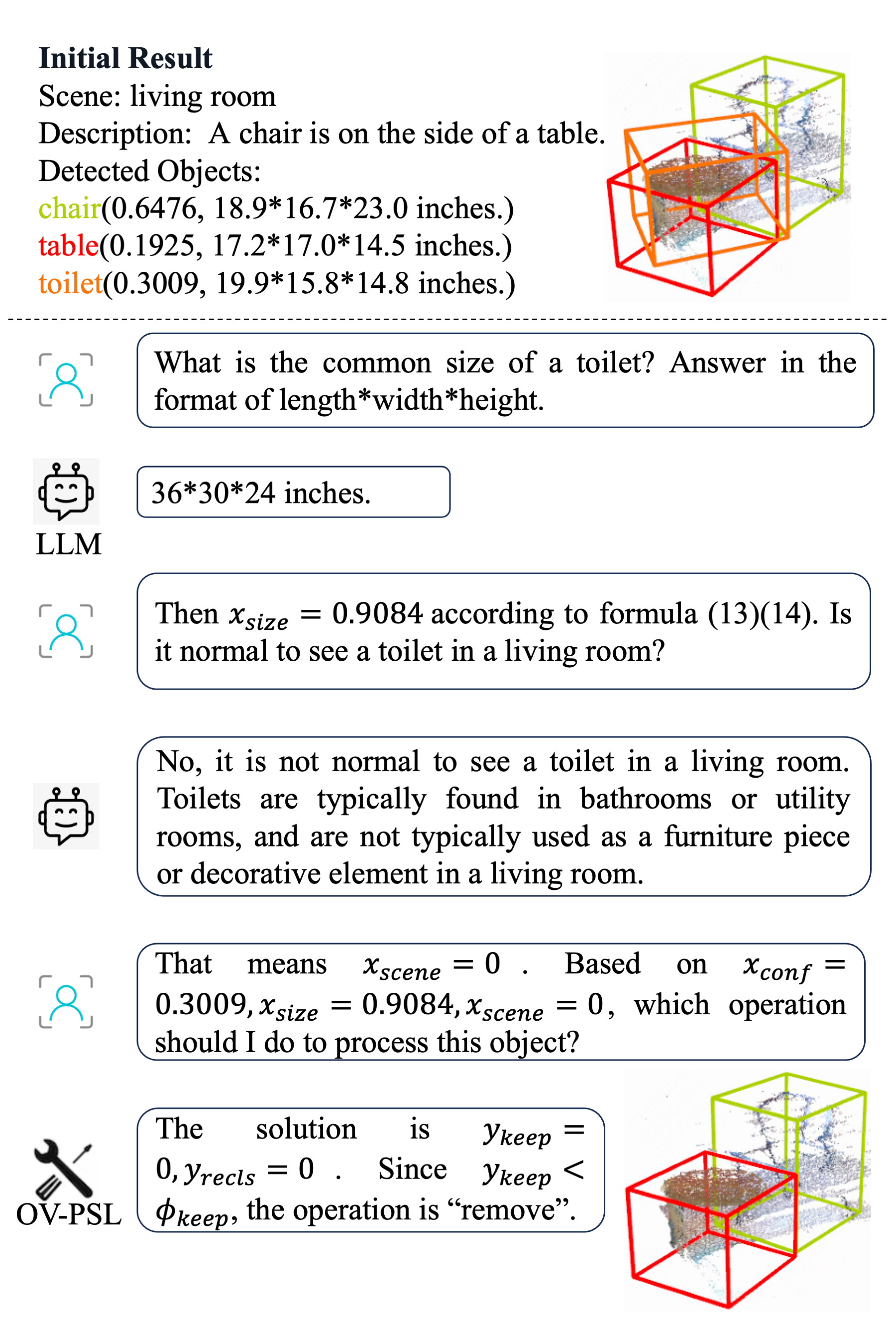}
}
\caption{Visualization of GLRD. In a living room, a toilet object is detected. The LLM recognizes that a toilet is unlikely to be placed in the living room, resulting in $x_{scene}=0$. Besides, the size of this object does not completely fit the common toilet, as $x_{size}=0.9084$. Taking all constraints into consideration, OV-PSL decides to remove this object.}
\label{viz1}
\end{figure}

\begin{figure*}[!t]
\centering
\resizebox{\textwidth}{!}{
\includegraphics{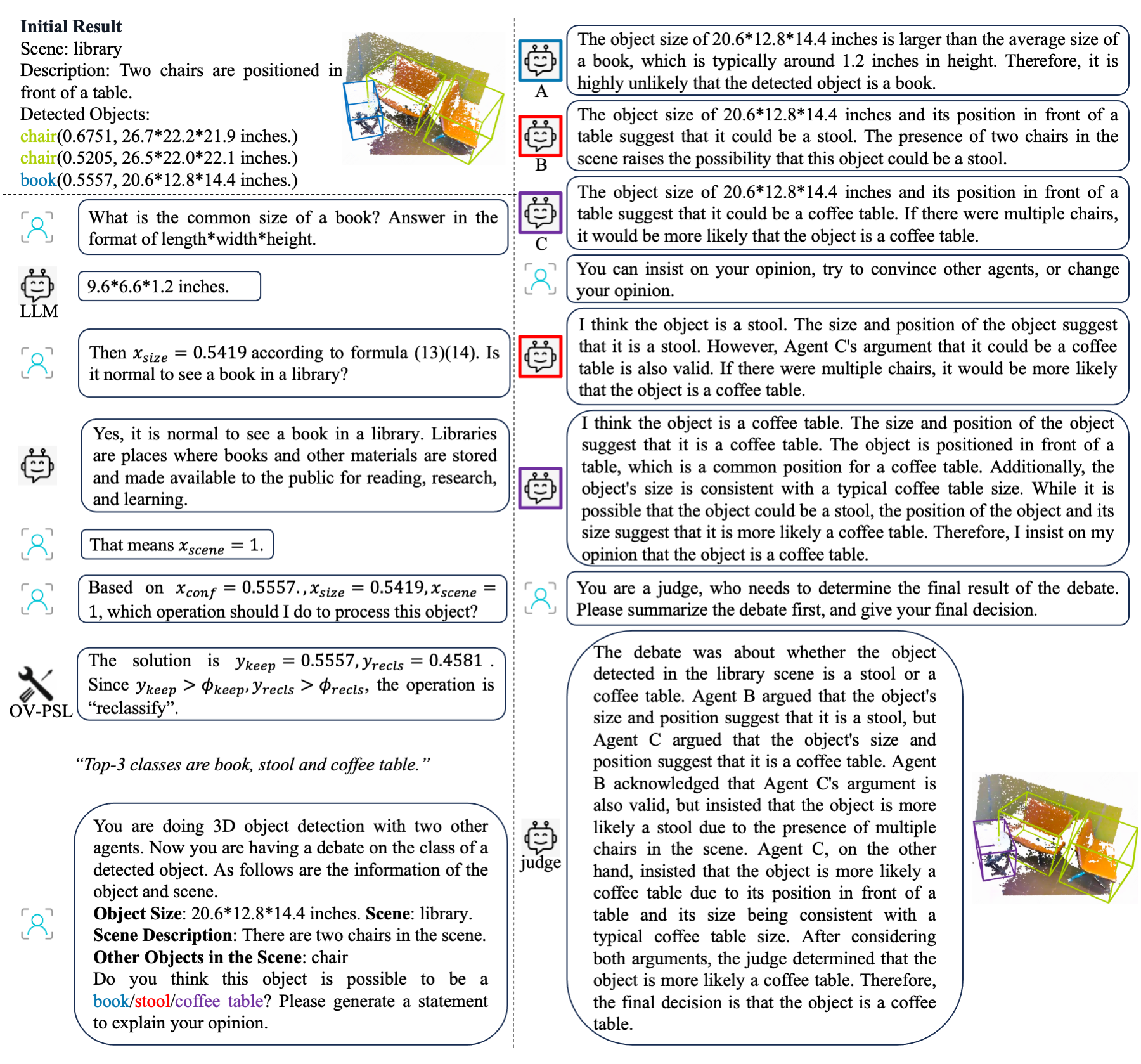}
}
\caption{Visualization of GLRD. In a library, two chairs and a book are detected. However, the size of the detected book is too large compared to a common book, indicating that this object is misclassified. Taking all constraints into consideration, OV-PSL decides to reclassify this object. The top-three classes with highest classification scores are set as candidate classes, each is represented by an agent. Through the debate of three agents, the object is finally classified as a coffee table by the judge.}
\label{viz2}
\end{figure*}

\textbf{\textit{Global-Local Collaboration:}} Global-Local Collaboration (without OV-PSL) greatly improves the detection performance of novel classes, whose average precision are improved by $0.83\%$. Note that we set $y_{keep}=\frac{1}{3}(x_{conf}+x_{size}+x_{scene}),y_{recls}=1-y_{keep}$ to avoid the use of OV-PSL. Although OV-PSL is not used, the collaboration still improves detection performance by a large gap, demonstrating the effect of the proposed global-local collaboration. Besides, we further analyze the parameters in the size error function (14), as shown in Tab. \ref{abla_alpha} and \ref{abla_size}. Specifically, $\alpha$ controls the importance of the size constraint, and low $\alpha$ values (\eg,$\alpha=0.05,0.15$) cannot fully release the effect of the size constraint. When $\alpha$ ranges from $0.25$ to $0.35$, the collaboration maintains a good performance, showing its robustness. However, too large $\alpha$ is harmful (\eg, $\alpha=0.45$), as the size constraint is overemphasized. For $\phi_{size}$, the detection performance is worsened when $\phi_{size}=0$, showing that the size constraint is too strict. When $\phi_{size}$ ranges from $0.03$ to $0.05$, the detection precisions maintain high, showing the robustness of the size constraint. However, too loose size constraints (\eg, $\phi_{size}=0.07,0.09$) can weaken the effect of Global-Local Collaboration.

\textbf{\textit{Probabilistic Soft Logic Solver for 3D OVD (OV-PSL):}} OV-PSL further stimulates Global-Local Collaboration to achieve better detection performance. For example, with OV-PSL, $AP_{25}^{novel}$ is improved by $1.18\%$, and $AP_{25}^{mean}$ is improved by $0.92\%$. Such results demonstrate that OV-PSL can effectively model the complex relationships between the common sense constraints and the operation scores. 

\subsection{Qualitative Analysis}

We present detailed examples of GLRD in Fig. \ref{viz1} and Fig. \ref{viz2}, showcasing its ability to refine the detection result through sophisticated reasoning processes.

In Fig. \ref{viz1}, the 3D detector initially identifies an object as a toilet within a living room. This detection triggers a series of reasoning steps, leveraging the LLM's understanding of common spatial arrangements. Specifically, the LLM recognizes that the presence of a toilet in a living room is atypical and inconsistent with usual household layouts, leading it to set the scene constraint $x_{scene}=0$. This indicates a high likelihood of a detection error based on contextual anomalies. Furthermore, LLM provides the size of a common toilet. Based on this information, GLRD evaluates the physical dimension of the detected object, determining that its size does not completely align with that of a standard toilet, resulting in a size confidence score of $x_{size}=0.9084$. Consequently, OV-PSL synthesize these constraints to conclude that the object should be removed from the detection result, as its presence and size do not match the common sense of indoor layouts.

In Fig. \ref{viz2}, the scene shows a library environment where a book and two chairs are detected. However, compared to a normal book, the size of the detected book is too large. Specifically, the size constraint $x_{size}=0.5419$, showing a high possibility that this object is misclassified. This discrepancy motivates OV-PSL to consider reclassifying the object. According to classification scores of this object, the top-three classes are book, stool and coffee table, which are set as candidate classes. In this way, GLRD conducts a debate involving three agents, each representing one of the candidate classes. These agents engage in a deliberation process, weighing the evidence and contextual cues to reach a consensus. Through this collaborative inference, the object is ultimately reclassified as a coffee table by the judge, a decision that aligns better with its size and the surrounding library context.

\section{Conclusion}
We propose a Global-Local Collaborative Reason and Debate with PSL (GLRD) framework for the 3D Open-Vocabulary Detection task. We propose RPLG to generate high-quality pseudo labels for training, and design BAOL to help the model distinguish foreground objects from background noise. To balance the model's attention towards different classes, we propose SBC for the balance of pseudo labels and DBC for the training loss. Besides, we propose Global-Local Collaboration to aggregate information from local and global branches for decisions, which refines the detection result. To unlock the potential of Global-Local Collaboration, we design OV-PSL to automatically solve optimal decisions. Extensive experiments on ScanNet and SUN RGB-D demonstrate the effect of GLRD. 

\bibliographystyle{splncs04}
\bibliography{egbib}

\vfill

\end{document}